\documentclass{article}

\usepackage[preprint]{nips_2018}

\usepackage[utf8]{inputenc} 
\usepackage[T1]{fontenc}    
\usepackage{hyperref}       
\usepackage{url}            
\usepackage{booktabs}       
\usepackage{amsfonts}       
\usepackage{nicefrac}       
\usepackage{microtype}      
\usepackage{graphicx}
\usepackage{amsmath}
\usepackage{amssymb}
\usepackage{multirow}
\usepackage{array}

\title{A Capsule-unified Framewok of Deep Neural Networks for Graphical Programming}

\author{
  Yujian Li\\
  College of Computer Science\\
  Faculty of Information Technology\\
  Beijing University of Technology\\
  Beijing, China 100124 \\
  \texttt{liyujian@bjut.edu.cn} \\
  \And
  Chuanhui Shan \\
  College of Computer Science\\
  Faculty of Information Technology\\
  Beijing University of Technology\\
  Beijing, China 100124 \\
  \texttt{chuanhuishan@emails.bjut.edu.cn} \\
}

\begin{document}

\maketitle

\begin{abstract}
 Recently, the growth of deep learning has produced a large number of deep neural networks. How to describe these networks unifiedly is becoming an important issue. We first formalize neural networks in a mathematical definition, give their directed graph representations, and prove a generation theorem about the induced networks of connected directed acyclic graphs. Then, using the concept of capsule to extend neural networks, we set up a capsule-unified framework for deep learning, including a mathematical definition of capsules, an induced model for capsule networks and a universal backpropagation algorithm for training them. Finally, we discuss potential applications of the framework to graphical programming with standard graphical symbols of capsules, neurons, and connections.
\end{abstract}

\section{Introduction}

Artificial neural networks are mathematical models that are inspired by biological neural networks. As a new era of neural networks, deep learning has become a powerful technology for artificial intelligence. Since presented by Hinton et al. [1], it has made a great amount of successes in image classification, speech recognition, and natural language processing [2-5], influencing on both academia and industry dramatically.

Essentially, deep learning is a collection of various methods for effectively training neural networks with deep structures. A neural network is usually regarded as a hierarchical system composed of many nonlinear computing units (or neurons, nodes). For example, a single-neuron network is the MP model proposed by McCulloch and Pitts in 1943 [6], with the function of performing logical operations. Although the MP model is unable to learn, it started the age of neural networks. In 1949, Hebb first proposed the idea of learning about biological neural networks [7]. In 1958, Rosenblatt invented a model of perceptron and its learning algorithm [8]. Subsequently, Minsky's criticism of the perceptron brought the study of neural networks to a period of low tide [9], but not stagnating completely. By the 1980s and 1990s, a worldwide research of neural networks came up with the presentation of Hopfield neural network [10], Boltzmann machine [11], multilayer perceptron [12], and so on.

\begin{figure}[htb]
  \centering
  \includegraphics[width=3in,height=1.7in]{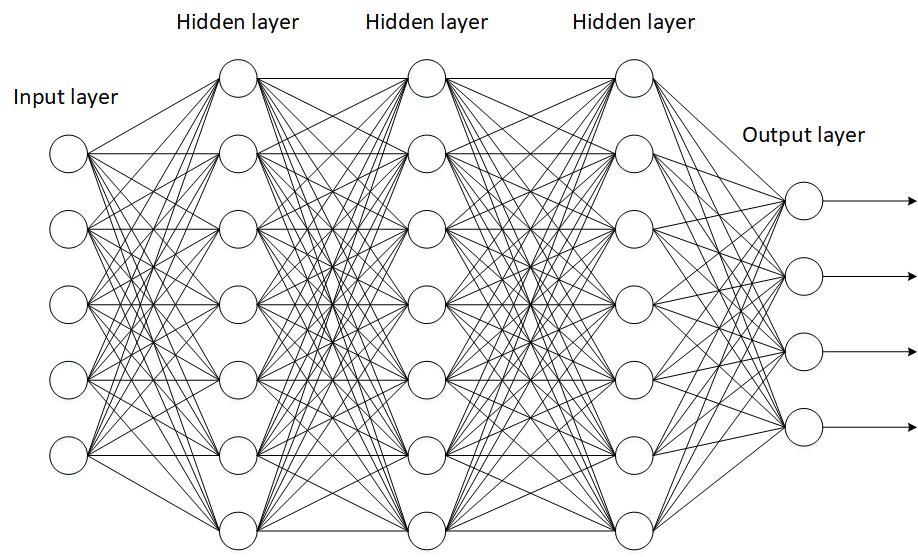}
  \caption{The structure of a MLP.}
\end{figure}

Multilayer perceptron (MLP) was once the most popular model of neural networks. A MLP consists of an input layer, a number of hidden layers and an output layer, as shown in Figure 1. The depth of it is the number of layers excluding the input layer. If the depth is greater than 2, a neural network may be called “deep”. For training MLPs, backpropagation (BP) is certainly the most well-known algorithm in common use [12, 13]. However, a majority of applications using MLPs were implemented with few hidden layers, while gaining little empirical benefit from additional hidden layers. Certain solace might be found in universal approximation theorems [14, 15], stating that a single-layer perceptron can approximate any multivariate continuous function arbitrarily well with enough hidden units. BP seemed to work only for shallow networks, albeit principally for deep networks. Not until 1991 was this problem on deep learning fully understood.

In 1991, Hochreiter made a milestone of deep learning [16], to formally indicate that typical deep networks suffer from the problem of vanishing or exploding gradients. The problem says, cumulative backpropagated error signals decay or explode exponentially in the number of layers, with consequence of shrinking rapidly or growing out of bounds. This is the major reason why BP is hard to train deep networks. Note that it could also be encountered as the long time lag problem in recurrent neural networks [17].

To overcome the training difficulties in deep neural networks (DNNs), in 2006 Hinton et al. started the new field of deep learning with a two-stage strategy in two landmark papers [1, 18], to demonstrate that unsupervised learning in shallow networks, e.g. contrastive divergence learning in restricted Boltzmann machines, can facilitate supervised learning in DNNs, e.g. BP in deep autoencoders (AEs) and deep MLPs. Furthermore, their pioneering work inspired many important techniques, including max pooling [19], dropout [20], dropconnect [21], batch normalization [22], etc. Additionally, a wealthy of historic accomplishments have been completed in applications such as image classification [2], speech recognition [3, 4], and natural language processing [5]. As a consequence, deep learning brings about a new and striking wave of neural networks in both academia and industry.

Broadly speaking, deep neural networks include feedforward neural networks (FNNs) and recurrent neural networks (RNNs). In this paper, we will focus on FNNs, for example, MLPs. Another kind of FNNs is convolutional neural networks (CNNs) that share connections and weights locally. CNNs were developed on the basis of neocognitrons [23, 24]. In 1998, LeCun et al. combined convolutional layers with downsampling layers to design an early structure of modern CNNs, namely, LeNet [25]. In 2012, Krizhevsky et al. used rectified linear units (ReLUs) and graphics processing units (GPUs) to build a breakthrough model of CNNs, i.e. AlexNet [2], which won the 2012 ImageNet Large Scale Visual Recognition Challenge. Since then, CNNs get into a rapid growth full of brilliant achievements and great contributions in the development of deep learning. They not only take the lead in competitions of image classification and recognition as well as object localization and detection [26-31], but also in a variety of application systems such as deep Q-networks [32], AlphaGo [33], speech recognition [34], and machine translation [5].

\begin{figure}[htb]
  \centering
  \includegraphics[width=4.5in,height=1.2in]{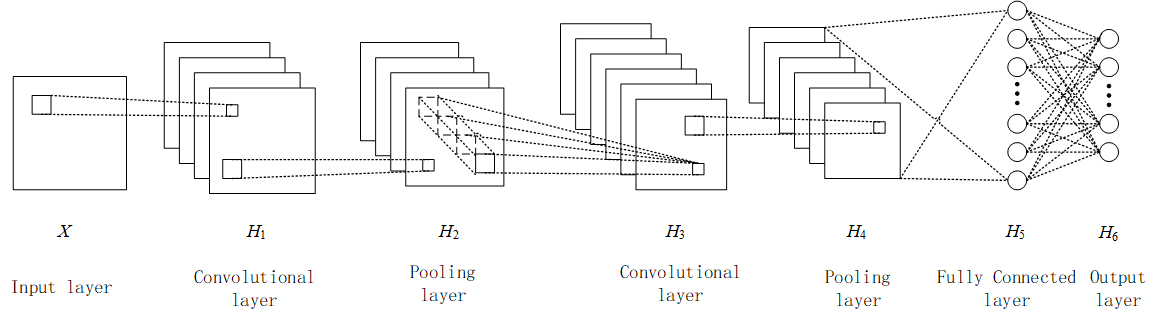}
  \caption{Structure of a CNN.}
\end{figure}

Theoretically, a CNN can be regarded as a special MLP. It may consist of an input layer, alternating convolutional and pooling layers, a fully connected layer, and an output layer, as shown in Figure 2. In fact, convolutional layers (or "detection layers"), and pooling layers (or “downsampling layers”), are special hidden layers. A convolutional kernel means the weight of a convolutional layer. Besides LeNet and AlexNet, many variants of CNNs have been presented, such as VGGNet [26], GoogLeNet [29], SqueezeNet [35], SPPNet [36], ResNet [27], DenseNet [37], FCN [38], Faster R-CNN [39], Mask R-CNN [40], YOLO [41], SSD [42], and 3D CNN [43]. There have been so many models of deep neural networks with different structures, even containing shortcut connections, parallel connections, and nested structures. The problem of how to establish a unified framework for DNNs, is becoming a progressively important issue.

In 2017 and 2018, Hinton et al. published two papers on capsule networks [44, 45], to overcome the disadvantages of CNNs. Different from their motivation, we will use capsules to establish a unified framework from a novel point of view. The unified framework has something to do with computational graphs [46]. A computational graph visualizes data dependence relations as an acyclic graph [47]. It may be composed of input vertices, non-input vertices, and dependence edges. Input vertices stand for input variables, non-input vertices stand for operations or transformations. A dependence edge stands for a directed relation between two vertices with one preceding the other. In Figure 3, input vertices are labelled by "$x_1$", "$x_2$" and "$x_3$", non-input vertices are labelled by "+", "$-$", "$\times$", and "/", and "$\sin$". The vertex "$\times$" has two preceding vertices "+" and "$-$". The dependence edges are all linear relations. For example, the directed edge from "$x_1$" to "+" has a weight of 3, meaning that $3{x_1}$ will be input into the vertex "+". It can be seen that $x_4=(2x_2)/(3x_3)$ is the output of vertex "/", $x_5 = -x_1-2x_4$ is the output of vertex "$-$", $x_6=3x_1+x_2$ is the output of vertex "+", $x_7=2x_6\times 0.5x_5$ is the output of vertex "$\times$", and $x_8 = \sin(0.3x_7)$ is the output of vertex "$\sin$". If $x_1=2$, $x_2=0$ and $x_3=3$, we have $x_4=0$, $x_5=-2$, $x_6=6$, $x_7 = -12$, $x_8 = \sin(-3.6)$.
\begin{figure}[htb]
  \centering
  \includegraphics[width=2.9in,height=1.3in]{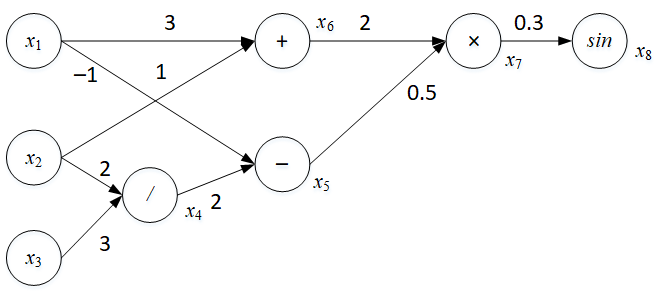}
  \caption{A computational graph that visualizes data dependence relations.}
\end{figure}

Note that a computational graph may not be a neural network. But a neural network can be viewed as a special computational graph. Like a computational graph, a neural network may also be composed of input vertices, non-input vertices, and dependences edges (or directed connections). However, non-input vertices of a neural network are only chosen from a special set of transformations, which are generally non-linear functions, such as sigmoid, tanh and ReLU. Usually, these transformations are termed activation functions. The input of each activation function (or node) is taken as the weighted sum of the outputs of its preceding nodes and a bias.

This paper is an extension of [48]. The remainder is organized as follows. In Section 2, we propose a mathematical definition to formalize neural networks, give their directed graph representations, and prove a generation theorem about the induced networks of connected directed acyclic graphs. In Section 3, we use the concept of capsule to extend neural networks, define an induced model for capsule networks, and establish a unified framework for deep learning with a universal backpropagation algorithm. In Section 4, we discuss potential applications of the unified framework to graphical programming. Finally, we make conclusions with future work in Section 5.

\section{Formalization of Neural networks}

\subsection{Mathematical definition}

A neural network is a computational model composed of neurons (or nodes) and connections. Nodes are divided into input neurons and non-input neurons. Input neurons represent real variables, e.g. $x_{1},x_{2},\cdots,x_{n}$. A non-input neuron can receive signals through connections both from input neurons and the outputs of other non-input neurons, and perform a weighted sum of these signals following a transformation termed activation function.

Let $X=\{x_{1},x_{2},\cdots,x_{n}\}$ stand for a set of real variables, and $F$ for a set of activation functions. On $X$ and $F$, a neural network can be formally defined as a 4-tuple $net=(S,H,W,Y)$. The 4-tuple is termed a neural network if it can be recursively generated by the following four rules.

1) \textbf{Rule of variable}. For any $z\in X$, let $y_z = z$. Let $S=\{z\}$, $H=\emptyset$, $W=\emptyset$, $Y=\{y_z\}$. Then $net=(S,H,W,Y)$ is a neural network.

2) \textbf{Rule of neuron}. For any nonempty subset $S\subseteq X$, $\forall f \in F$, $\forall b \in \mathbb{R}$, construct a non-input neuron $h\not\in X$ that depends on $(f,b)$, select weighting connections $w_{x_i\rightarrow h} (x_i\in S)$, and set $y_h = f(\sum_{x_i \in S} {w_{x_i\rightarrow h} x_i}+b)$. Let $H=\{h\}$, $W=\{w_{x_i\rightarrow h} |x_i\in S\}$, and $Y=\{y_h\}$. Then, $net=(S,H,W,Y)$ is a neural network.

3) \textbf{Rule of growth}. Suppose $net=(S,H,W,Y)$ is a neural network. For any nonempty subset $N\subseteq S\cup H$, $\forall f\in F$, $\forall b\in \mathbb{R}$, construct a non-input neuron $h\not\in S\cup H$ that depends on $(f,b)$, select weighting connections  $w_{z_j\rightarrow h}(z_j\in N)$, and set $y_h = f(\sum_{z_j\in N} {w_{z_j\rightarrow h}y_{z_j}}+b)$. Let $S'=S$, $H'=H\cup \{h\}$, $W'=W\cup \{w_{z_j\rightarrow h}|z_j \in N\}$, and $Y'=Y\cup \{y_h\}$. Then, $net'=(S',H',W',Y')$ is also a neural network.

4) \textbf{Rule of convergence}. Suppose $net_k=(S_k,H_k,W_k,Y_k)(1\leq k \leq K)$ are $K$ neural networks, satisfying that $\forall 1\leq i\neq j \leq K$, $(S_i \cup H_i) \cap (S_j\cup H_j)=\emptyset$. For any nonempty subsets $A_k \subseteq S_k \cup H_k(1\leq k \leq K)$, $N=\bigcup_{k=1}^K {A_k}$, $\forall f\in F$, $\forall b \in \mathbb{R}$, construct a non-input neuron $h\not\in \bigcup_{k=1}^K (S_k\cup H_k)$ that depends on $(f,b)$, select weighting connections $w_{z\rightarrow h}(z\in N)$, and set $y_h=f(\sum_{z\in N} {w_{z\rightarrow h}y_z}+b)$. Let $S=\bigcup_{k=1}^K S_k$, $H=(\bigcup_{k=1}^K H_k)\cup \{h\}$, $W=(\bigcup_{k=1}^K W_k)\cup \{w_{z\rightarrow h}|z\in N\}$, and $Y=(\bigcup_{k=1}^K Y_k)\cup \{y_h\}$, then $net=(S,H,W,Y)$ is also a neural network.

In the above four rules, weighting connection $w_{z\rightarrow h}$ is a data structure that contain the full information of the directed connection $z\rightarrow h$, including its strength, back-end neuron $z$ and front-end neuron $h$. Moreover, if a neuron $h$ depends on $(f,b)$, $f$ is called the activation function of $h$, and $b$ is called the bias of $h$. Additionally, if the 4-tuple $net=(S,H,W,Y)$ is a neural network, $S$ is called the set of input neurons, $H$ the set of non-input neurons, $W$ the set of weighting connections, and $Y$ the set of outputs. Finally, the rule of neuron can be deduced from the rule of variable and the rule of convergence. In fact, for any nonempty subset $S=\{x_{i_1},x_{i_2},\cdots,x_{i_r}\}\subseteq X$, we can use the rule of variable to generate $r$ neural networks $net_{r'}=(\{x_{r'}\},\emptyset,\emptyset,\{x_{r'}\})$ $1\leq r'\leq r$. According to the rule of convergence, we can also use the $r$ networks to generate a new neural network $net=(S,H,W,Y)$, where $S=\bigcup_{r'=1}^r S_{r'}$, $H= \{h\}$, $W=\{w_{z\rightarrow h}|z\in S\}$, and $Y=\{y_h|y_h=f(\sum_{z \in S} {w_{z \rightarrow h} y_z}+b)\}$. This is exactly the rule of neuron.
\begin{figure}[htb]
  \centering
  \includegraphics[width=2.3in,height=0.5in]{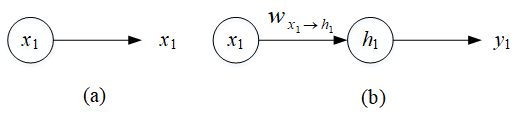}
  \caption{(a) A trivial network; (b) A 1-1 network}
\end{figure}

\begin{figure}[htb]
  \centering
  \includegraphics[width=4.7in,height=0.9in]{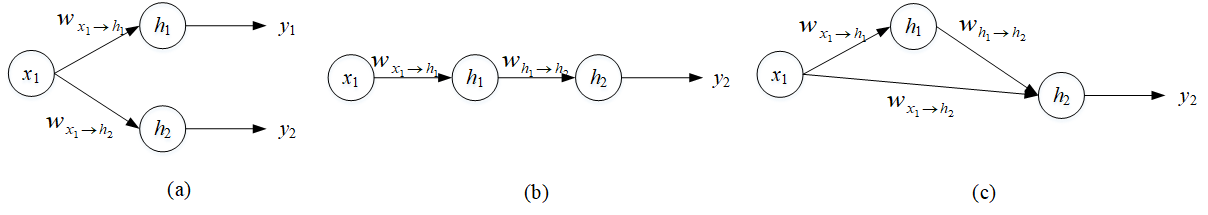}
  \caption{Three 1-2 networks}
\end{figure}

\subsection{Directed graph representation}

Let $X$ be a set of real variables and $F$ be a set of activation functions. For any neural network $net=(S,H,W,Y)$ on $X$ and $F$, a directed acyclic graph $G_{net}=(V,E)$ can be constructed with the set of vertices $V=S\cup H$ and the set of directed edges $E=\{z\rightarrow h|w_{z\rightarrow h} \in W\}$. $G_{net}=(V,E)$ is called the directed graph representation of $net=(S,H,W,Y)$. How to construct such a graph representation is detailed in the following two cases.

\begin{figure}[htb]
  \centering
  \includegraphics[width=5in,height=3.5in]{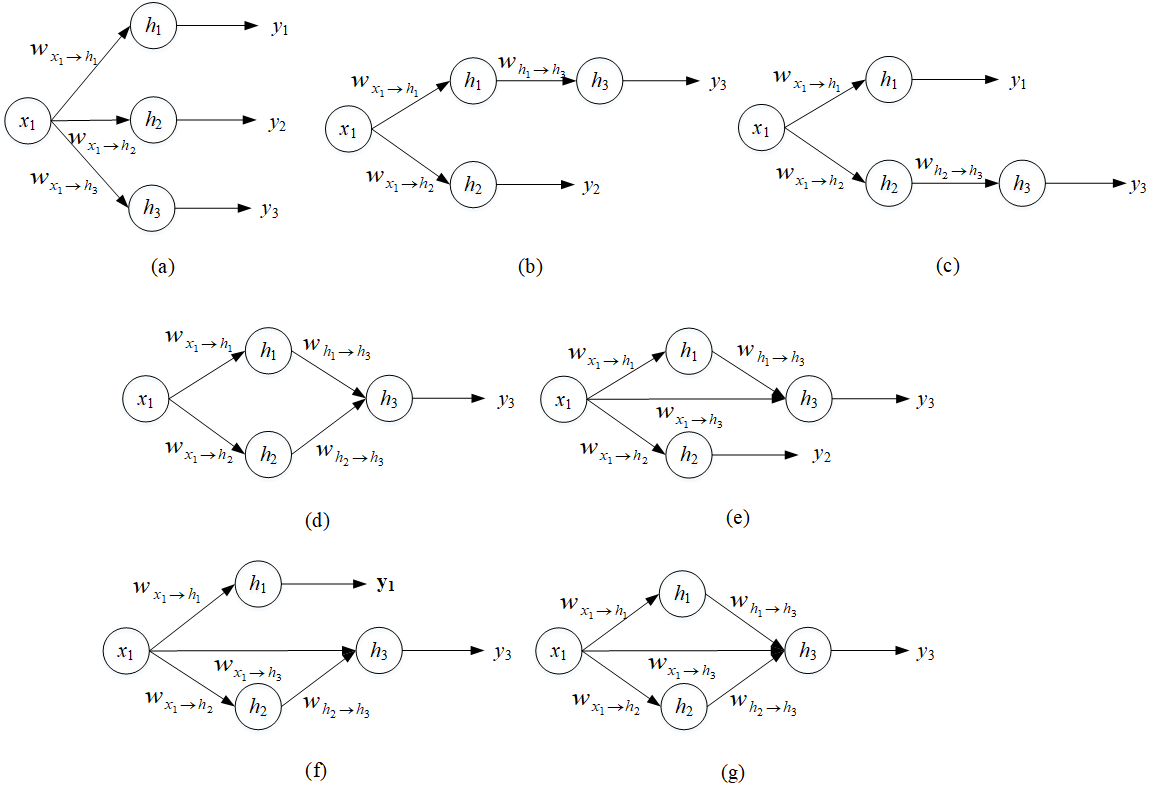}
  \caption{Seven 1-3 networks}
\end{figure}

\textbf{1)The case of $X=\{x_1\}$}

For $x_1 \in X$, let $y_{x_1}=x_1$, $S=\{x_1\}$, $H=\emptyset$, $W=\emptyset$, $Y=\{y_{x_1}\}$. According to the rule of variable, $net=(S,H,W,Y)$ is a neural network, as shown in Figure 4(a). This network is called "trivial network". It has only one input neuron, which does nothing.

For a nonempty subset $S=\{x_1\} \subseteq X$, $\forall f\in F$, $\forall b \in \mathbb{R}$, construct a non-input neuron $h_1 \not\in S$ that depends on $(f,b)$, select weighting connection $w_{x_1\rightarrow h_1}$, and set $y_{h_1}=f(w_{x_1\rightarrow h_1}x_1 +b)$. Let $H=\{h_1\}$, $W=\{w_{x_1\rightarrow h_1}\}$, and $Y=\{y_{h_1}\}$. According to the rule of neuron, $net=(S,H,W,Y)$ is a neural network, as shown in Figure 4(b). This network is called "1-1 network". It has one input neuron and one non-input neuron.

Using the rule of growth on the 1-1 network, three different neural networks can be generated next, as shown in Figures 5(a-c). These networks are called "1-2 networks". Each of them has one input neuron and two non-input neurons.

Using the rule of growth on the three 1-2 networks, twenty-one different neural networks can be totally generated further. They are all 1-3 networks, with seven displayed in Figures 6(a-g). The seven are generated from the network in Figure 5(a).

\textbf{2)The case of $X=\{x_1,x_2\}$}

For $x_1, x_2\in X$, let $y_{x_1}=x_1$, $y_{x_2}=x_2$, $S_1=\{x_1\}$, $S_2=\{x_2\}$, $H_1=H_2=\emptyset$, $W_1=W_2=\emptyset$, $Y_1=\{y_{x_1}\}$ and $Y_2=\{y_{x_2}\}$. Using the rule of variable, $net_1=(\{x_1\},\emptyset,\emptyset,\{y_{x_1}\})$ and $net_2=(\{x_2\},\emptyset,\emptyset,\{y_{x_2}\})$ are neural networks. Obviously, both of them are trivial networks.
\begin{figure}[htb]
  \centering
  \includegraphics[width=1.2in,height=0.6in]{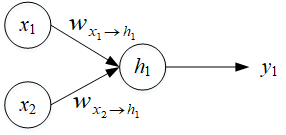}
  \caption{A 2-1 network.}
\end{figure}

\begin{figure}[htb]
  \centering
  \includegraphics[width=5.5in,height=1.8in]{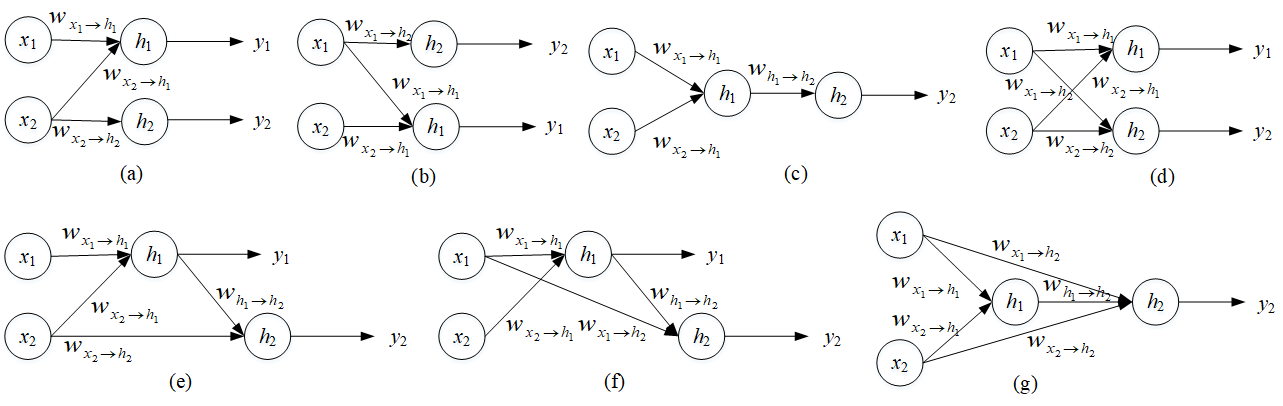}
  \caption{Seven 2-2 networks.}
\end{figure}

If $S=\{x_1,x_2\}$, $\forall f\in F$, $\forall b \in \mathbb{R}$, construct a non-input neuron $h_1\not\in S$ that depends on $(f,b)$, select weighting connections $w_{x_i\rightarrow h_1}(x_i \in S)$, and set $y_{h_1}=f(\sum_{x_i\in S}{w_{x_i\rightarrow h_1}x_i}+b)$. Let $H=\{h_1\}$, $W=\{w_{x_1\rightarrow h_1}, w_{x_2\rightarrow h_1}\}$, and $Y=\{y_{h_1}\}$. According to the rule of neuron, $net=(S,H,W,Y)$ is a neural network. This is a 2-1 network, as depicted in Figure 7. Using the rule of growth on the 2-1 network, seven different 2-2 networks can be generated next, as depicted in Figures 8(a-g).
\begin{figure}[htb]
  \centering
  \includegraphics[width=5.5in,height=6.5in]{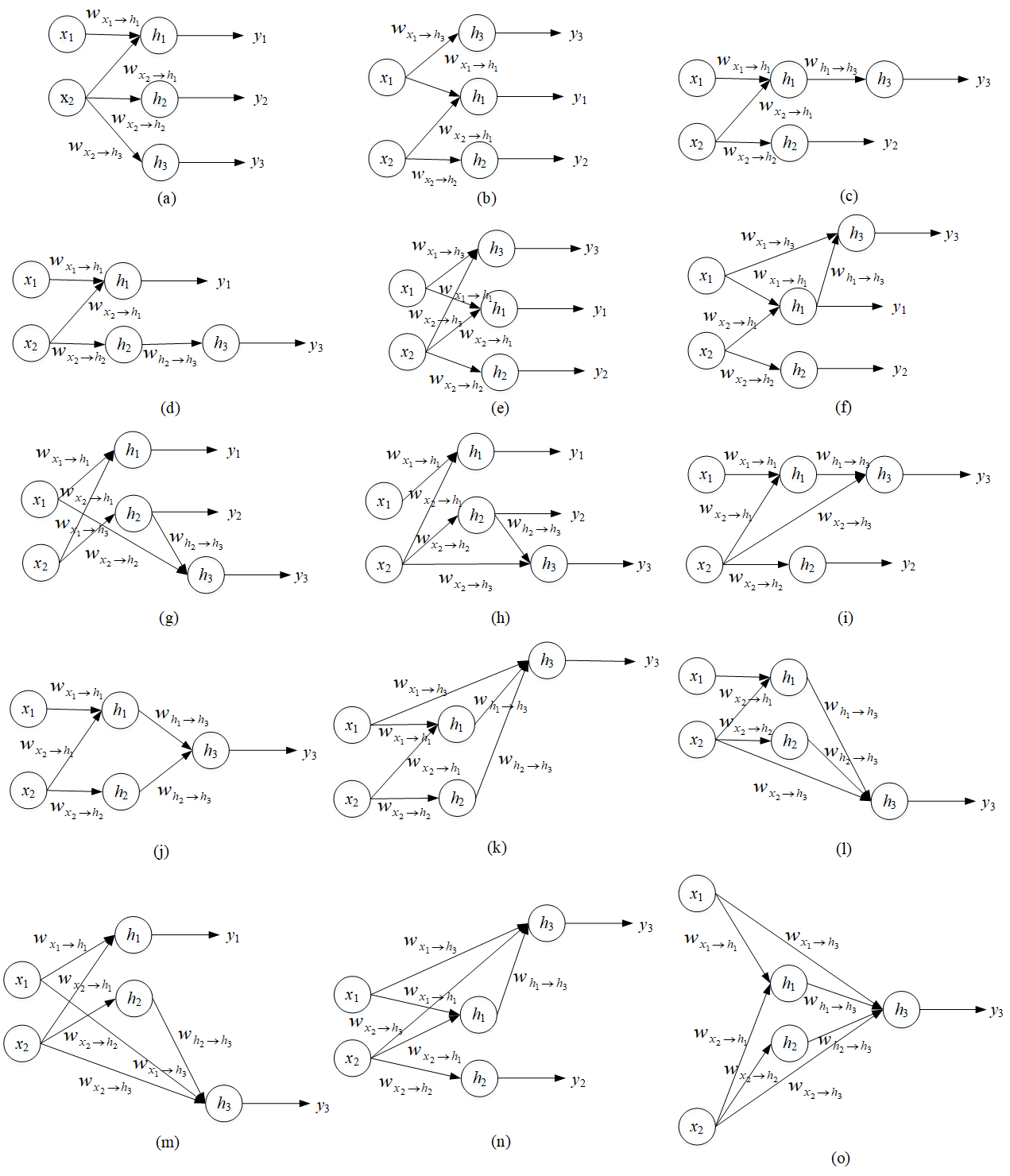}
  \caption{Fifteen 2-3 networks.}
\end{figure}

Using the rule of growth on the seven networks in Figure 8, one hundred and five different neural networks can be totally generated further, with fifteen shown in Figures 9(a-o). The fifteen are all 2-3 networks, generated from the network in Figure 8(a).
\begin{figure}[htb]
  \centering
  \includegraphics[width=4.3in,height=0.9in]{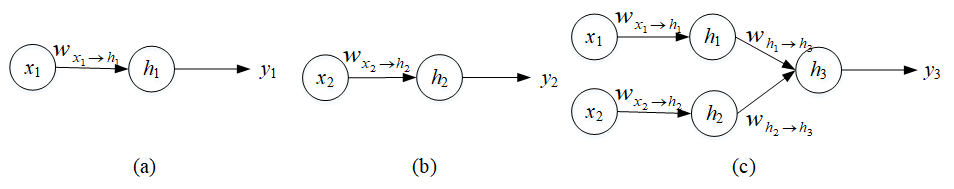}
  \caption{A necessary explanation for the rule of convergence.}
\end{figure}

Finally, the rule of convergence is necessary. In fact, it cannot generate all neural networks only using the three rules of variable, neuron and growth. An example is the network in Figure 10(c). Without the rule of convergence, this network cannot be generated from the two in Figures 10(a-b).

\subsection{Induced network and its generation theorem}

Suppose $G=(V,E)$ is a connected directed acyclic graph, where $V$ is the set vertices (nodes) and $E$ is the set of directed edges. For any vertex $h\in V$, let ${IN}_h=\{z|z\in V, z\rightarrow h \in E\}$ be the set of all vertices that precede $h$ adjacently. Let ${OUT}_h=\{z|z\in V,h\rightarrow z \in E\}$ be the set of vertices that follow $h$ adjacently. If $IN_h=\emptyset$, then $h$ is called an input node of $G$. If $OUT_h=\emptyset$, then $h$ is called an output node of $G$. Otherwise, $h$ is called a hidden node of $G$. Let $X$ stand for the set of all input nodes, $O$ for the set of all output nodes, and $M$ for the set of all hidden nodes. Obviously, $V=X\cup M\cup O$, and $M=V-X\cup O$.

Let $y_h$ be the output of node $h$, and $w_{z\rightarrow h}$ be the weighting connection from $z$ to $h$. Then, a computational model of graph $G$ is defined as follows:

\hspace{0.4cm}1) $\forall z \in X$, $y_z=z$.

\hspace{0.4cm}2) $\forall h\in M\cup O$, select $f\in F$ and $b \in \mathbb{R}$ to compute $y_h=f(\sum_{z\in IN_h}{w_{z\rightarrow h}y_z}+b)$.

Let $S=X$, $H=M\cup O$, $W=\{w_{z\rightarrow h}|z\rightarrow h \in E\}$, and $Y=\{y_h|h\in V\}$. Then, $net_G = (S,H,W,Y)$ is called an induced network of graph $G$. We have the following generation theorem.

\textbf{Generation Theorem:} For any connected directed acyclic graph $G=(V,E)$, its induced network $net_G$ is a neural network that can be recursively generated by the rules of variable, neuron, growth, and convergence.

\textbf{Proof:} By induction on $|V|$ (i.e. number of vertices), we prove the theorem as follows.\\
1) When $|V|=1$, we have $|X|=1$ and $|O|=0$, so the induced network $net_G$ is a neural network that can be generated directly by the rule of variable.\\
2) When $|V|=2$, we have $|X|=1$ and $|O|=1$, so the induced network $net_G$ is a neural network that can be generated directly by the rule of growth.\\
3) Assume that the theorem holds for $|V|\leq n$. When $|V|= n+1\geq 3$, the induced network $net_G$ has at least one output node $h\in O$. Let $E_h=\{z\rightarrow h\in E\}$ denote the set of edges heading to the node $h$. Moreover, let $V'=V-\{h\}$ and $E'=E-E_h$. Based on the connectedness of $G'=(V',E')$, we have two cases to discuss in the following:
\begin{enumerate}
\item[i)] If $G'=(V',E')$ is connected, then applying the induction assumption for $|V'|\leq n$, the induced network $net_{G'}=(S',H',W',Y')$ can be recursively generated by the rules of variable, neuron, growth, and convergence. Let $N=IN_h$. In $net_G=(S,H,W,Y)$, we use $f\in F$ and $b\in \mathbb{R}$ to stand for the activation function and bias of node $h$, and $w_{z\rightarrow h}(z\in N)$ for the weighting connection from node $z$ to the node $h$. Then, $net_G$ can be obtained by using the rule of growth on $net_{G'}$, to generate the node $h$ and its output $y_h = f(\sum_{z\in N}{w_{z\rightarrow h}y_z}+b)$.
\item[ii)] Otherwise, $G'$ comprises a number of disjoint connected components $G_k=(V_k,E_k)(1\leq k \leq K)$. Using the induction assumption for $|V_k|\leq n(1\leq k \leq K)$, the induced network $net_{G_k}=(S_k,H_k,W_k,Y_k)$ can be recursively generated by the rules of variable, neuron, growth, and convergence. Let $A_k=(S_k\cup H_k)\cap IN_h$, and $N= \bigcup_{k=1}^K{A_k}$. In $net_G=(S,H,W,Y)$, we use $f\in F$ and $b\in \mathbb{R}$ to stand for the activation function and bias of the node $h$, and $w_{z\rightarrow h}(z\in N)$ for the weighting connection from node $z$ to node $h$. Then, $net_G$ can be obtained by using the rule of convergence on $net_{G_k}(1\leq k \leq K)$, to generate the node $h$ and its output $y_h = f(\sum_{z\in N}{w_{z\rightarrow h}y_z}+b)$.
\end{enumerate}
As a result, the theorem always holds.

Note that in the generation theorem, the rule of neuron can be deleted, because it can be deduced from the rule of variable and the rule of convergence.

\section{Capsule framework of Deep learning}

\subsection{Mathematical definition of capsules}

In 2017, Hinton et al. pioneered the idea of capsules and considered a nonlinear "squashing" capsule [45], as shown in Figure 11. The capsule has $n$ input vectors $\textbf{u}_1,\textbf{u}_2,\cdots,\textbf{u}_n$, and $n$ weight matrices $\textbf{w}_1,\textbf{w}_2,\cdots,\textbf{w}_n$. Let $\textbf{\^{u}}_i=\textbf{w}_i\textbf{u}_i(1\leq i \leq n)$ be prediction vectors. The total input is $\textbf{s}=\sum_i c_i\textbf{\^{u}}_i=\sum_i c_i\textbf{w}_i\textbf{u}_i$, where $c_i$ is a coupling coefficient between low-level capsule $i$ and the capsule itself. The value of $c_i$ is determined by a dynamic routing process, satisfying $\sum_i c_i =1$. The output of the capsule $\textbf{v}$ is a vector computed by the nonlinear squashing function
\begin{equation}
  \textbf{v}=squash(\textbf{s})=\frac{\|\textbf{s}\|^2}{1+\|\textbf{s}\|^2}\frac{\textbf{s}}{\|\textbf{s}\|}.
\end{equation}

\begin{figure}[htb]
  \centering
  \includegraphics[width=3in,height=0.9in]{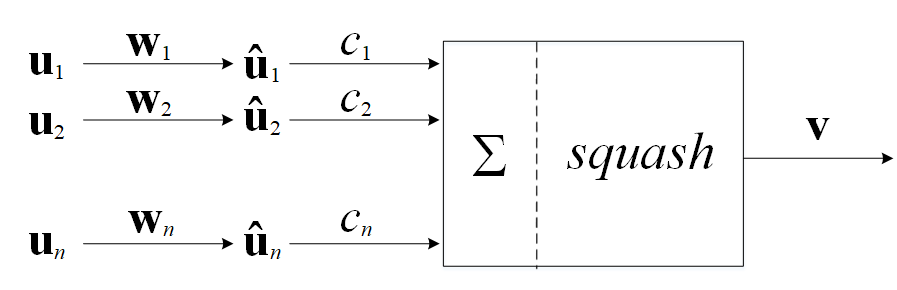}
  \caption{A model of squashing capsule.}
\end{figure}
\begin{figure}[htb]
  \centering
  \includegraphics[width=2.7in,height=1.0in]{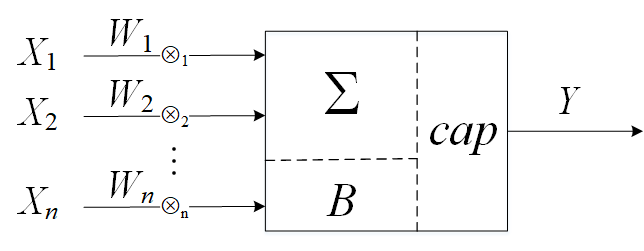}
  \caption{A model of general capsule.}
\end{figure}

From the viewpoint of mathematical models, a capsule is essentially an extension of traditional activation functions. A traditional activation function has a scalar input and a scalar output. However, a capsule can have a vector input and a vector output. More generally, a capsule may have a tensor input and a tensor output. As shown in Figure 12, a capsule may have $n$ input tensors $X_1,X_2,\cdots,X_n$, $n$ weight tensors $W_1,W_2,\cdots,W_n$, and a bias $B$. In addition, $\otimes_1,\otimes_2,\cdots,\otimes_n$ are $n$ weighting operations, which may be taken as "identity transfer", "scalar multiplication", "dot product", "matrix multiplication", "convolution", and so on. Meanwhile, $W_i\otimes_iX_i(1\leq i \leq n)$ and $B$ must be tensors with the same dimension. The total input of the capsule is $U=\sum_i{W_i\otimes_iX_i}+B$, and the output $Y$ is a tensor computed by a nonlinear capsule function $cap$, namely,
\begin{equation}
  Y=cap(U)=cap(\sum_i{W_i\otimes_iX_i}+B).
\end{equation}

According to Eq. (2), a capsule is a generalization of scalar activation function with scalar inputs.

\subsection{Definition of Capsule Networks}

For convenience, we use $\mathcal{F}$ to stand for a nonempty set of capsule functions, and $\mathbb{T}$ for the set of all tensors.

Suppose $\mathcal{G}=(\mathcal{V},\mathcal{E})$ is a connected directed acyclic graph, where $\mathcal{V}$ denotes the set of vertices and $\mathcal{E}$ denotes the set of directed edges. For any vertex $H\in \mathcal{V}$, let $IN_H$ be the set of all vertices that precede $H$ adjacently. Let $OUT_H$ be the set of vertices that follow $H$ adjacently. If $IN_H=\emptyset$, then $H$ is called an input vertex of $\mathcal{G}$. If $OUT_H=\emptyset$, then $H$ is called an output vertex of $\mathcal{G}$. Otherwise, $H$ is called a hidden vertex of $\mathcal{G}$. Let $\mathcal{X}$ stand for the set of all input vertices, $\mathcal{O}$ for the set of all output vertices, and $\mathcal{M}$ for the set of all hidden vertices. Obviously, $\mathcal{V}=\mathcal{X}\cup \mathcal{M}\cup \mathcal{O}$, and $\mathcal{M}=\mathcal{V}-\mathcal{X}\cup \mathcal{O}$.

Furthermore, let $Y_H$ be the output of vertex $H$, and $(W_{Z\rightarrow H},\otimes_{Z\rightarrow H})$ be the tensor-weighting connection from $Z$ to $H$. If $\forall H\in \mathcal{M}\cup \mathcal{O}$, $\forall Z \in IN_H$, $W_{Z\rightarrow H}\otimes_{Z\rightarrow H}Y_Z$ and $B$ are tensors with the same dimension, then a tensor-computational model of graph $\mathcal{G}$ is defined as follows:

\hspace{0.4cm}1) $\forall Z \in \mathcal{X}$, $Y_Z=Z$,

\hspace{0.4cm}2) $\forall H\in \mathcal{M}\cup \mathcal{O}$, select $cap\in \mathcal{F}$ and $B \in \mathbb{T}$ to compute $Y_H=cap(\sum_{Z\in IN_H}{W_{Z\rightarrow H}\otimes_{Z\rightarrow H}Y_Z}+B)$.

 Let $\mathcal{S}=\mathcal{X}$, $\mathcal{H}=\mathcal{M}\cup \mathcal{O}$, $\mathcal{W}=\{(W_{Z\rightarrow H},\otimes_{Z\rightarrow H})|Z\rightarrow H\in \mathcal{E}\}$, and $\mathcal{Y}=\{Y_H|H\in \mathcal{V}\}$. Then, $net_{\mathcal{G}}=(\mathcal{S},\mathcal{H},\mathcal{W},\mathcal{Y})$ is called a tensor-induced network of graph $\mathcal{G}$. This network is also called a capsule network, where the vertices stand for capsules, and the directed edges for weighting connections. If $net_{\mathcal{G}}=(\mathcal{S},\mathcal{H},\mathcal{W},\mathcal{Y})$ is a capsule network, $\mathcal{S}$ is called the set of input capsules, $\mathcal{H}$ the set of non-input capsules, $\mathcal{W}$ the set of weighting connections, and $\mathcal{Y}$ the set of output tensors.

Using a capsule network, a MLP can be simplified as a directed acyclic path. For example, the MLP in Figure 1 can be represented as the capsule path in Figure 13. In fact, this MLP has five layers: one input layer, three hidden layers, and one output layer. On the whole, each layer can be thought of as a capsule. Let $X=(x_1,x_2,\cdots,x_5)^T$ stand for the input capsule, $H_i=(cap_i,B_i)(i=1,2,3)$ for the hidden capsules, and $O=(cap_4,B_4)$ for the output capsule. The capsule function $cap_i$ is defined by the elementwise activation function, and the capsule bias $B_i$ by the bias vector. If all the weighting operations $\otimes_{X\rightarrow H_1}$, $\otimes_{H_1\rightarrow H_2}$, $\otimes_{H_2\rightarrow H_3}$, and $\otimes_{H_3\rightarrow O}$ are all taken as matrix multiplication "$\times$", then we are easy to obtain $(W_{X\rightarrow H_1},\otimes_{X\rightarrow H_1})=((w_{m,n}^{X\rightarrow H_1})_{7\times5},\times)$, $(W_{H_1\rightarrow H_2},\otimes_{H_1\rightarrow H_2})=((w_{m,n}^{H_1\rightarrow H_2})_{7\times7},\times)$, $(W_{H_2\rightarrow H_3},\otimes_{H_2\rightarrow H_3})=((w_{m,n}^{H_2\rightarrow H_3})_{7\times7},\times)$ and $(W_{H_3\rightarrow O},\otimes_{H_3\rightarrow O})=((w_{m,n}^{H_3\rightarrow O})_{4\times7},\times)$, which are the tensor-weighting connections from $H_0=X$ to $H_1$, $H_1$ to $H_2$, $H_2$ to $H_3$ and $H_3$ to $O$. Finally, let $Y_{H_i}(i=1,2,3)$ stand for the output vector of $H_i$, and $Y_O$ for the output vector of $O$. Setting $Y_{H_O}=X$ and $Y_{H_4}=Y_O$, we have $Y_{H_i} = cap_i(W_{H_{i-1}\rightarrow H_i}\times Y_{H_{i-1}}+B_i)$ for $1\leq i \leq 4$. This means that the MLP in Figure 1 can be represented as the capsule path in Figure 13.
\begin{figure}[htb]
  \centering
  \includegraphics[width=3.5in,height=0.4in]{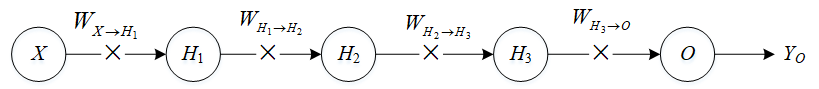}
  \caption{A capsule structure of MLP, with "$\times$" standing for matrix multiplication.}
\end{figure}

Besides MLPs, capsule networks can also be used to simplify the structures of other DNNs. For example, the CNN in Figure 2 can be represented as the capsule path in Figure 14. In fact, this CNN has 7 layers: one input layer, two convolutional layers, two downsampling (pooling) layers, one fully connected layer, and one output layer. On the whole, each of the layers can be thought of as a capsule. Let $X$ stand for the input capsule, $H_i=(cap_i,B_i)(i=1,\cdots,5)$ for the hidden capsules, and $O=(cap_6,B_6)$ for the output capsule. The capsule functions $cap_1$ and $cap_3$ are defined by the elementwise ReLU function. $cap_2$ and $cap_4$ are defined by downsampling "$\downarrow$". $cap_5$ is the identity function. $cap_6$ is the softmax function. Moreover, the capsule biases $B_i(i=1,\cdots,6)$ are each defined by the bias tensor of the corresponding layer. Let both $\otimes_{X\rightarrow H_1}$ and $\otimes_{H_2\rightarrow H_3}$ be convolution operation "$\ast$", both $\otimes_{H_1\rightarrow H_2}$ and $\otimes_{H_3\rightarrow H_4}$ be identity transfer "$\rightarrow $", $\otimes_{H_4\rightarrow H_5}$ be tensor-reshaping operation "$\triangleleft$", and $\otimes_{H_5\rightarrow O}$ be matrix multiplication "$\times$". Then, $(W_{X\rightarrow H_1},\otimes_{X\rightarrow H_1})=(W_{X\rightarrow H_1},\ast)$, $(W_{H_1\rightarrow H_2},\otimes_{H_1\rightarrow H_2})=("",\rightarrow)$, $(W_{H_2\rightarrow H_3},\otimes_{H_2\rightarrow H_3})=(W_{H_2\rightarrow H_3},\ast)$, $(W_{H_3\rightarrow H_4},\otimes_{H_3\rightarrow H_4})=("",\rightarrow)$, $(W_{H_4\rightarrow H_5},\otimes_{H_4\rightarrow H_5})=("",\triangleleft)$, and $(W_{H_5\rightarrow O},\otimes_{H_5\rightarrow O})=(W_{H_5\rightarrow O},\times)$, which are tensor-weighting connections from $X$ to $H_1$, $H_1$ to $H_2$, $H_2$ to $H_3$, $H_3$ to $H_4$, $H_4$ to $H_5$, and $H_5$ to $O$. Finally, we have
\begin{equation}
\begin{cases}
  Y_{H_1} = cap_1(W_{X\rightarrow H_1}\ast X+B_1) = \textrm{ReLU}(W_{X\rightarrow H_1}\ast X+B_1), \\
  Y_{H_2} = cap_2(\rightarrow Y_{H_1}+B_2) = \downarrow Y_{H_1}+B_2, \\
  Y_{H_3} = cap_3(W_{H_2\rightarrow H_3}\ast X+B_3) = \textrm{ReLU}(W_{H_2\rightarrow H_3}\ast Y_{H_2}+B_3), \\
  Y_{H_4} = cap_4(\rightarrow Y_{H_3}+B_4) = \downarrow Y_{H_3}+B_4, \\
  Y_{H_5} = cap_5(\triangleleft Y_{H_4}+B_5) = \triangleleft Y_{H_4}, \\
  Y_O = cap_6(W_{H_5\rightarrow O}\times Y_{H_5}+B_6) = softmax(W_{H_5\rightarrow O}\times Y_{H_5}+B_6).
\end{cases}
\end{equation}

According to Eq. (3), the CNN in Figure 2 can be represented as the capsule path in Figure 14.
\begin{figure}[htb]
  \centering
  \includegraphics[width=4.8in,height=0.4in]{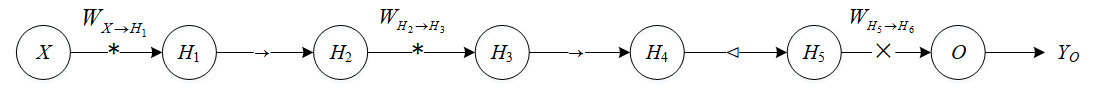}
  \caption{A Capsule structure of CNN, with "$\ast$" standing for convolution operation, "$\rightarrow$" for identity transfer, "$\triangleleft$" for tensor reshaping, and "$\times$" for matrix multiplication.}
\end{figure}

\subsection{Types of capsule networks}

As mentioned in Section 3.2, a MLP can be simplified as a path structure composed of capsules, so can many CNNs (e.g. LeNet and VGGNet). Note that the path structure is a special case of layered structure. Generally speaking, capsule networks can be divided into two types: layered capsule networks and skip capsule networks. A layer capsule network can only have weighting connections must between two adjacent layers (see Figure 15), whereas a skip capsule network may have weighting connections between any two layers (see Figure 16).
\begin{figure}[htb]
  \centering
  \includegraphics[width=4.0in,height=1.4in]{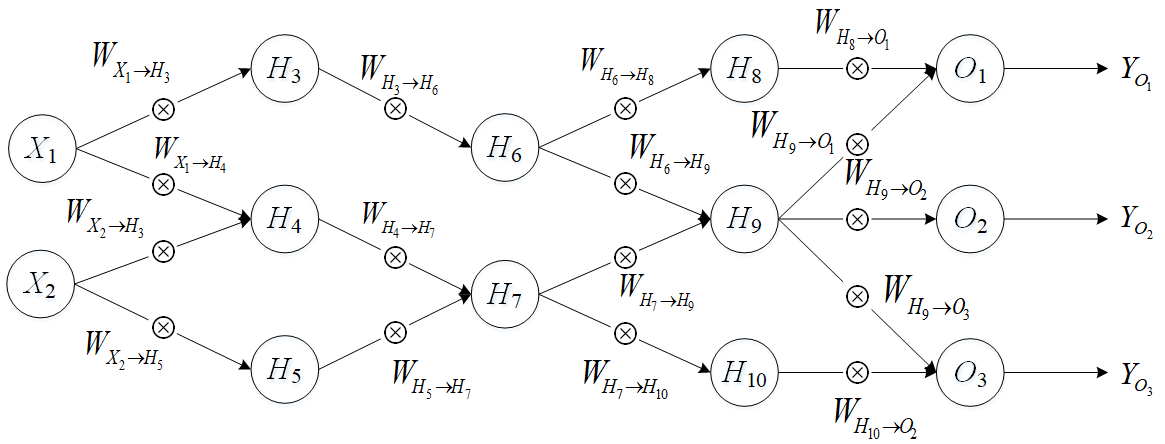}
  \caption{A layered capsule network with 5 layers.}
\end{figure}

\begin{figure}[htb]
  \centering
  \includegraphics[width=3.5in,height=1.8in]{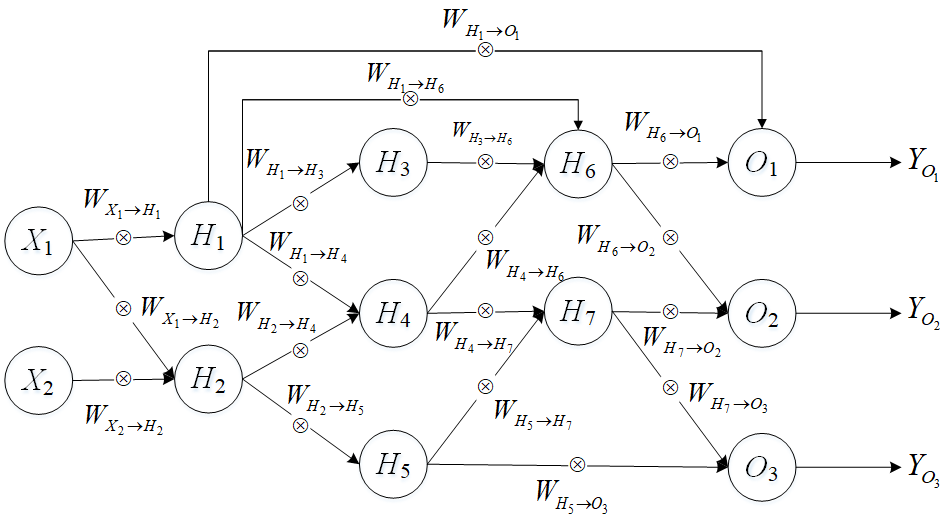}
  \caption{A skip capsule network.}
\end{figure}

Suppose $\mathcal{G}=(\mathcal{V},\mathcal{E})$ is a connected directed acyclic graph, and $net_{\mathcal{G}} = (\mathcal{S},\mathcal{H},\mathcal{W},\mathcal{Y})$ is a tensor-induced network of graph $\mathcal{G}$. We have $\mathcal{V}=\mathcal{S}\cup\mathcal{H}$ and $\mathcal{H}=\mathcal{M}\cup\mathcal{O}$. The capsule network $net_{\mathcal{G}}$ is called "layered", if the set of hidden capsules $\mathcal{M}$ can be partitioned into a number of disjoint subsets $\mathcal{M}_1,\mathcal{M}_2,\cdots,\mathcal{M}_k$, such that
\begin{enumerate}
  \item $\mathcal{M}=\mathcal{M}_1\cup\mathcal{M}_2\cup\cdots\cup\mathcal{M}_k$, $\mathcal{M}_0=\mathcal{S}$, $\mathcal{M}_{k+1}=\mathcal{O}$,
  \item $\forall H\rightarrow H'\in \mathcal{E}$, $\exists i\in \{0,1,\cdots,k\}$, $H\in \mathcal{M}_i$ and $H'\in \mathcal{M}_{i+1}$,
  \item $\forall H,H'\in \mathcal{M}_i(i=0,1,\cdots,k+1)$, $H\rightarrow H'\not\in \mathcal{E}$ and $H'\rightarrow H \not\in \mathcal{E}$.
\end{enumerate}

If $net_{\mathcal{G}}=(\mathcal{S},\mathcal{H},\mathcal{W},\mathcal{Y})$ is not "layered", it is called "skip".

In Figure 15, we are easy to have a layered struture: $\mathcal{M}_0=\{X_1,X_2\}$, $\mathcal{M}_1=\{H_3,H_4,H_5\}$, $\mathcal{M}_2=\{H_6,H_7\}$, $\mathcal{M}_3=\{H_8,H_9,H_{10}\}$ and $\mathcal{M}_4=\{O_1,O_2,O_3\}$. However, in Figure 16, we cannot have such a layered structure.

\subsection{Universal backpropagation of capsule networks}

A capsule network is a tensor- induced network of graph $\mathcal{G}=(\mathcal{V},\mathcal{E})$. It can be denoted as $net_{\mathcal{G}}=(\mathcal{S},\mathcal{H},\mathcal{W},\mathcal{Y})$. Let the set of input capsules $\mathcal{S}=\mathcal{X}=\{X_1,X_2,\cdots,X_n\}$. Suppose $\mathcal{O}=\{O_1,O_2,\cdots,O_m\}$ is the set of output capsules. Then, $\mathcal{H}=\mathcal{V}-\mathcal{S}$ is the set of non-input capsules, and $\mathcal{M}=\mathcal{V}-\mathcal{X}\cup \mathcal{O}=\{H_1,H_2,\cdots,H_l\}$ is the set of hidden capsules.

If the total number of capsules $|\mathcal{S}\cup \mathcal{H}|\geq 2$, then for $\forall H\in \mathcal{H}$, we have
\begin{equation}
\begin{cases}
  U_H = \sum_{Z\in IN_H}{W_{Z\rightarrow H}\otimes_{Z\rightarrow H}Y_Z+B_H}, \\
  Y_H = cap_H(U_H)=cap_H(\sum_{Z\in IN_H}{W_{Z\rightarrow H}\otimes_{Z\rightarrow H}Y_Z+B_H}).
\end{cases}
\end{equation}

For any output node $H\in \mathcal{O}$, let $Y_H$ and $T_H$ be its actual output tensor and expected output tensor respectively for the set of input capsules $\mathcal{S}= \mathcal{X}$. The loss function between the two tensors is defined as
\begin{equation}
  L_H=Loss(Y_H,T_H).
\end{equation}

Accordingly, the total loss function can be computed by
\begin{equation}
L=\sum_{H\in \mathcal{O}}L_H.
\end{equation}

Let $\delta_H=\frac{\partial L}{\partial U_H}$ denote the backpropagated error signal (or sensitivity) for capsule $H$. By the chain rule, we are easy to further obtain:
\begin{equation}
\forall H\in \mathcal{O},
\begin{cases}
  \delta_H & = \frac{\partial L}{\partial U_H} = \frac{\partial Loss(Y_H,T_H)}{\partial Y_H}\cdot \frac{\partial cap_H}{\partial U_H}, \\
  \frac{\partial L}{\partial B_H} & = \frac{\partial L}{\partial U_H}\cdot \frac{\partial U_H}{\partial B_H} = \delta_H, \\
  \frac{\partial L}{\partial W_{Z\rightarrow H}} & = \frac{\partial L}{\partial U_H}\cdot \frac{\partial U_H}{\partial W_{Z\rightarrow H}} =\delta_H \cdot \frac{\partial U_H}{\partial W_{Z\rightarrow H}}.
\end{cases}
\end{equation}

\begin{equation}
\forall H\in \mathcal{M},
\begin{cases}
  \delta_H & = \frac{\partial L}{\partial U_H} = \sum_{P\in OUT_H}{\frac{\partial L}{\partial U_P}\cdot \frac{\partial U_P}{\partial Y_H}\cdot \frac{\partial Y_H}{\partial U_H}} \\
& = \sum_{P\in OUT_H}{\delta_P \cdot \frac{\partial U_P}{\partial Y_H}\cdot \frac{\partial cap_H}{\partial U_H}}, \\
  \frac{\partial L}{\partial B_H} & = \frac{\partial L}{\partial U_H}\cdot \frac{\partial U_H}{\partial B_H}=\delta_H, \\
  \frac{\partial L}{\partial W_{Z\rightarrow H}} & = \frac{\partial L}{\partial U_H}\cdot \frac{\partial U_H}{\partial W_{Z\rightarrow H}}=\delta_H \cdot \frac{\partial U_H}{\partial W_{Z\rightarrow H}}.
\end{cases}
\end{equation}

It should be noted that in formulae (7)-(8), the computation of $\frac{\partial cap_H}{\partial U_H}$ depends on the specific form of capsule function $cap_H$. For example, when   is an elementwise sigmoid function, the result is:
\begin{equation}
\frac{\partial cap_H}{\partial U_H}=sigmoid(U_H)(1-sigmoid(U_H)).
\end{equation}

Additionally, $\frac{\partial U_H}{\partial W_{Z\rightarrow H}}$ and $\frac{\partial U_P}{\partial Y_H}$ also depends on the specific choice of weighting operation $\otimes_{Z\rightarrow H}$.

Based on formulae (7)-(8), a universal backpropagation algorithm can be designed for capsule networks, with one iteration detailed in \textbf{Algorithm 1}. This algorithm can have many variants from different versions of gradient descent [49].
\begin{table*}
  \centering
  \begin{tabular}{lll}
    \multicolumn{3}{c}{\textbf{Algorithm 1}: One iteration of the universal backpropagation algorithm.} \\
    \toprule
    1) Select a learning rate $\eta > 0$, \\
    2) $\forall H\in \mathcal{M}\cup \mathcal{O}$, $\forall Z \in IN_H$, initialize $W_{Z\rightarrow H}$ and $B_H$,  \\
    3) $\forall H\in \mathcal{O}$, compute $\delta_H = \frac{\partial Loss(Y_H,T_H)}{\partial Y_H}\cdot \frac{\partial cap_H}{\partial U_H}$, \\
    4) $\forall H \in \mathcal{M}$, compute $\delta_H = \sum_{P\in OUT_H}{\delta_P \cdot \frac{\partial U_P}{\partial Y_H}\cdot \frac{\partial cap_H}{\partial U_H}}$, \\
    5) Compute $\Delta W_{Z\rightarrow H}=\delta_H\cdot \frac{\partial U_H}{\partial W_{Z\rightarrow H}}$ and $\Delta B_H = \delta_H$, \\
    6) Update $W_{Z\rightarrow H} \leftarrow W_{Z\rightarrow H}- \eta \cdot \Delta W_{Z\rightarrow H}$,$B_H \leftarrow B_H- \eta \cdot \Delta B_H$.\\
    \bottomrule
  \end{tabular}
\end{table*}

\begin{table*}
  \centering
  \begin{tabular}{lll}
    \multicolumn{3}{c}{\textbf{Algorithm 2}: A universal backpropagation algorithm on training pairs in many iterations.} \\
    \toprule
    Input: training pairs $(\mathcal{X}^k,\mathcal{T}^k)(k=1,\cdots,K)$, \\
    Output: $W_{Z\rightarrow H}$, $\Delta W_{Z\rightarrow H}$, $B_H$, $\Delta B_H$, \\
    Select a learning rate $\eta > 0$, \\
    $\forall H\in \mathcal{M}\cup \mathcal{O}$, $\forall Z \in IN_H$, initialize $W_{Z\rightarrow H}$ and $B_H$,  \\
    \textbf{for} iter=1:max\_iter \textbf{do} \\
        \hspace{0.4cm} \textbf{for} $\forall H\in \mathcal{O}$, \textbf{do} $\delta_H^k = \frac{\partial Loss(Y_H^k,T_H^k)}{\partial Y_H^k}\cdot \frac{\partial cap_H^k}{\partial U_H^k}$ \textbf{end for} \\
        \hspace{0.4cm} \textbf{for} $\forall H \in \mathcal{M}$, \textbf{do} $\delta_H^k = \sum_{P\in OUT_H^k}{\delta_P^k \cdot \frac{\partial U_P^k}{\partial Y_H^k}\cdot \frac{\partial cap_H^k}{\partial U_H^k}}$ \textbf{end for} \\
        \hspace{0.4cm} $\Delta W_{Z\rightarrow H}=\sum_{k=1}^K {\delta_H^k \cdot \frac{\partial U_H^k}{\partial W_{Z\rightarrow H}}}$ and $\Delta B_H = \sum_{k=1}^K \delta_H^k$ \\
        \hspace{0.4cm} $W_{Z\rightarrow H} \leftarrow W_{Z\rightarrow H}- \eta \cdot \Delta W_{Z\rightarrow H}$,$B_H \leftarrow B_H- \eta \cdot \Delta B_H$\\
    \textbf{end for} \\
    \bottomrule
  \end{tabular}
\end{table*}

In practice, we need a training algorithm on pairs $(\mathcal{X}^k,\mathcal{T}^k)(k=1,\cdots,K)$, where $\mathcal{X}^k=\{X_1^k,X_2^k,\cdots,X_n^k\}$ and $\mathcal{T}^k=\{T_1^k,T_2^k,\cdots,T_n^k\}$. For any $H\in \mathcal{O}$, we have the loss between $Y_H^k$ and $T_H^k$,
\begin{equation}
  L_H^k=Loss(Y_H^k,T_H^k).
\end{equation}

and the total loss for the $k$-th pair:
\begin{equation}
  L^k=\sum_{H\in \mathcal{O}}L_H^k.
\end{equation}

Therefore, we can compute the total loss for all training pairs as follows
\begin{equation}
  L=\sum_{k=1}^K L^k=\sum_{k=1}^K {\sum_{H\in \mathcal{O}}L_H^k}.
\end{equation}

Using gradient descent on formula (12), we can have a universal backpropagation algorithm on training pairs in many iterations, which is detailed in \textbf{Algorithm 2}.

\section{Potential Applications to Graphical Programming}

The capsule framework can be potentially applied to designing a graphical programming platform based on a set of standard graphical symbols for capsule networks and plain networks. Using this platform, we can implement deep neural networks by drawing them directly instead of coding. The platform is not only expected to make easier deep learning, but also to promote its development significantly.

\subsection{Standard graphical symbols for capsule networks}

In future applications, the capsule framework will be able to provide a theoretical basis for graphical programming on deep learning. Under the framework, we can design a set of standard graphical symbols to draw capsule networks, mainly including capsule symbols and connection symbols.
\begin{figure}[htb]
  \centering
  \includegraphics[width=3.2in,height=0.5in]{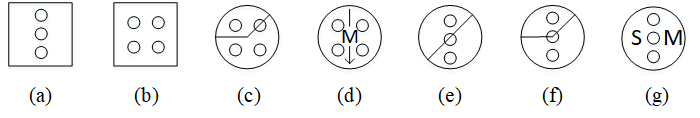}
  \caption{Examples of standard capsule symbols: (a) 1D-data capsule, (b) 2D-data capsule, (c) 2D-ReLU capsule, (d) 2D-maximum downsampling capsule, (e) 1D-identical capsule, (f) 1D-ReLU capsule, (g) 1D-softmax capsule.}
\end{figure}

\begin{table}[h]
  \centering
  \caption{Types and attributes of some standard capsule symbols.}
  \begin{tabular}{|p{0.4\columnwidth}<{\centering}|p{0.4\columnwidth}<{\centering}|}\hline
    Capsule type & Attributes \\ \hline
    1D-data capsule & (dimension $M$, data type $T$) \\ \hline
    2D-data capsule & (height $M$, width $N$, channel number $d$, data type $T$) \\ \hline
    2D-ReLU capsule & (height $M$, width $N$, channel number $d$, data type $T$) \\ \hline
    2D-maximum downsampling capsule & (height $M$, width $N$, channel number $d$, downsampling window $\lambda \times \tau$, data type $T$) \\ \hline
    1D-identical capsule & (dimension $M$, data type $T$) \\ \hline
    1D-ReLU capsule & (dimension $M$, data type $T$) \\ \hline
    1D-softmax capsule & (dimension $M$, data type $T$) \\ \hline
  \end{tabular}
\end{table}

Some of standard capsule symbols are shown in Figure 17 (a-g), such as 1D-data capsule, 2D-data capsule, 2D-ReLU capsule, 2D-maximum downsampling capsule, 1D-identical capsule, 1D-ReLU capsule, and 1D-softmax capsule. Their types and attributes are listed in Table 1, with more detail given as follows.

1) 1D-data capsule is a structure of vector array for the input of a capsule network. It can be represented as $\mathcal{X}=X=(x_1,\cdots,x_M)$, with 2 basic attributes: dimension $M$ and data type $T$ (e.g. float64).

2) A 2D-data capsule is a structure of matrix array for the input of a capsule network. It can be represented as $\mathcal{X}=(X_1,\cdots,X_d)$, with 4 basic attributes: height $M$, width $N$, channel number $d$ and data type $T$. Moreover, each $X_i(1\leq i \leq d)$ is a matrix of size $M\times N$ and data type $T$.

3) A 2D-ReLU capsule transforms a matrix array input into another matrix array output with elementwise ReLU. It has four basic attributes: height $M$, width $N$, channel number $d$ and data type $T$. Its input is a matrix array $\mathcal{X}=(X_1,\cdots,X_d)$, where each $X_i$ is of size $M\times N$ and data type $T$. Its output is also a matrix array $\mathcal{Y}=(Y_1,\cdots,Y_d)$, defined as:
\begin{equation}
\mathcal{Y}=ReLU(\mathcal{X})=(ReLU(X_1),\cdots,ReLU(X_d)).
\end{equation}

4) A 2D-maximum downsampling capsule transforms a matrix array input into another matrix array output with maximum downsampling function $\downarrow_{max}^{\lambda,\tau}$. It has five basic attributes: height $M$, width $N$,  channel number $d$, downsamplinged window $\lambda \times \tau$ and data type $T$. Its input is a matrix array $\mathcal{X}=(X_1,\cdots,X_d)$, where each $X_i$ is of size $M\times N$ and data type $T$. Its output is also a matrix array $\mathcal{Y}=(Y_1,\cdots,Y_d)$, defined as:
\begin{equation}
\mathcal{Y}=\downarrow_{max}^{\lambda,\tau}(\mathcal{X})=(\downarrow_{max}^{\lambda,\tau}(X_1),\cdots,\downarrow_{max}^{\lambda,\tau}(X_d)),
\end{equation}
\begin{equation}
Y_i=\downarrow_{max}^{\lambda,\tau}(X_i)=(\downarrow_{max}^{\lambda,\tau}(G_{\lambda,\tau}^{X_i})).
\end{equation}
where $G_{\lambda,\tau}^{A}$ denotes the non-overlapping block matrix of $A$ divided by $\lambda \times \tau$. If $A$ is a $M\times N$ matrix, then $G_{\lambda,\tau}^{A}$ is a block matrix that contains $\frac{M}{\lambda}$ rows and $\frac{N}{\tau}$ columns. The $i,j$-th block can be expressed as
\begin{equation}
G_{\lambda,\tau}^{A}(i,j) = (a_{st})_{\lambda \times \tau}, (i-1)\times \lambda+1\leq s \leq i \times \lambda, (j-1)\times \tau+1\leq t \leq j \times \tau.
\end{equation}
The maximum downsampling of $G_{\lambda,\tau}^{A}(i,j)$ is defined as
\begin{equation}
\downarrow_{max}^{\lambda,\tau}(G_{\lambda,\tau}^{A}(i,j)) = \max\{a_{st}|(i-1)\times \lambda+1\leq s \leq i \times \lambda, (j-1)\times \tau+1\leq t \leq j \times \tau\}.
\end{equation}
The maximum downsampling of $G_{\lambda,\tau}^{A}$ is defined as
\begin{equation}
\downarrow_{max}^{\lambda,\tau}G_{\lambda,\tau}^{A} = (\downarrow_{max}^{\lambda,\tau}(G_{\lambda,\tau}^{A}(i,j))).
\end{equation}

5) A 1D-identical capsule transforms a vector input into another vector output with the identical function $I$. It has two basic attributes: dimension $M$ and data type $T$. Its input is a vector $\mathcal{X}=(X)=X$ of dimension $M$ and data type $T$. Its output is also a vector $\mathcal{Y}=Y$, defined as:
\begin{equation}
Y=I(X)=X.
\end{equation}

6) A 1D-ReLU capsule transforms a vector input into another vector output with elementwise ReLU. It has two basic attributes: dimension $M$ and data type $T$. Its input is a vector $\mathcal{X}=X$ of dimension $M$ and data type $T$. Its output is a vector $\mathcal{Y}=Y$, defined as:
\begin{equation}
Y=ReLU(X).
\end{equation}

7) A 1D-softmax capsule transforms a vector input into another vector output with function $softmax$. It has two basic attributes: dimension $M$ and data type $T$. Its input is a vector $\mathcal{X}=X$ of dimension $M$ and data type $T$. Its output is a vector $\mathcal{Y}=Y$, defined as:
\begin{equation}
Y=softmax(X).
\end{equation}

\begin{figure}[htb]
  \centering
  \includegraphics[width=4in,height=0.5in]{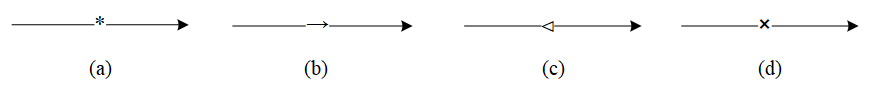}
  \caption{Examples of standard connection symbols: (a) convolutional connection, (b) transfer connection, (c) reshaping connection, (d) full connection.}
\end{figure}

\begin{table}[h]
  \centering
  \caption{Types and attributes of several types of standard connected graphical symbols.}
  \begin{tabular}{|p{0.18\columnwidth}<{\centering}|p{0.22\columnwidth}<{\centering}|p{0.22\columnwidth}<{\centering}|p{0.22\columnwidth}<{\centering}|}
    \hline
    \multirow{2}*{Connection types} & \multicolumn{3}{c|}{Attributes} \\ \cline{2-4}
                                    & Back-end structure & Weight structure & Front-end structure \\ \hline
    Convolutional connection & (height $M$, width $N$, channel number $d$, data type $T$) & (kernel number $k$, height $m$, width $n$, channel number $d$, stride $s$, data type $T$) & (height $M-m+1$, width $N-n+1$, channel number $k$, data type $T$) \\ \hline
    Transfer connection & (height $M$, width $N$, channel number $d$, data type $T$) & (height $M$, width $M$, data type $T$) & (height $M$, width $N$, channel number $d$, data type $T$) \\ \hline
    Reshaping connection & (height $M$, width $N$, channel number $d$, data type $T$) & (dimension $M$, channel number $M$, data type $T$) & (dimension $dMN$, data type $T$) \\ \hline
    Full connection & (dimension $N$, data type $T$) & (height $M$, width $N$, data type $T$) & (dimension $N$, data type $T$) \\ \hline
  \end{tabular}
\end{table}

Some of standard connection symbols are shown in Figure 18, such as "convolutional connection", "transfer connection", "reshaping connection", and "full connection". Their types and attributes are listed in Table 2 and further explained as follows.

1) A convolutional connection transforms the back-end capsule into the front-end capsule with convolution operation. It has three attributes: back-end structure, weight structure and front-end structure. The back-end structure is a matrix array $\mathcal{X}=(X_1,\cdots,X_d)$, where each $X_i$ is of size $M\times N$ and data type $T$. he weight structure is a $k$-kernel tensor array $\mathcal{W}=(\mathcal{W}_1,\cdots,\mathcal{W}_k)$ with $\mathcal{W}_i(1\leq i\leq k)$ representing the $i$-th convolutional kernel. Moreover, the tensor $\mathcal{W}_i = (W_{1}^{i},\cdots,W_{d}^{i})$ is a $d$-channel matrix array, where each $W_{j}^{i}$ of size $m\times n$ and data type $T$. The front-end structure $\mathcal{Y}=(Y_1,\cdots,Y_k)$ is also a matrix array, defined as:
\begin{equation}
\mathcal{Y}=\mathcal{W}*\mathcal{X}=(\mathcal{W}_1,\cdots,\mathcal{W}_k)*\mathcal{X}=(\mathcal{W}_1*\mathcal{X},\cdots,\mathcal{W}_k*\mathcal{X}),
\end{equation}
where
\begin{equation}
Y_i=\mathcal{W}_i*\mathcal{X}=\sum_{j=1}^{d}W_j^i*X_j.
\end{equation}
Note that $W_j^i * X_j$ is calculated by matrix convolution. If $A_{m\times n}$ and $B_{M\times N}$ are two matrices, then their $s$-stride convolution $C = (c_{ij})=A_{m\times n}*B_{M\times N}$ is a matrix of size $(\frac{M-m}{s}+1)\times (\frac{N-n}{s}+1)$ and data type $T$, with $c_{ij}$ defined as:
\begin{equation}
c_{ij}= \sum_{u=1}^{m}\sum_{v=1}^{n}{a_{uv}\cdot b_{is+m-u-1,js+n-v-1}}, 1\leq i \leq \frac{M-m}{s}+1, 1\leq j \leq \frac{N-n}{s}+1.
\end{equation}

2) A transfer connection transforms the back end into the front end with identity operation. It has three attributes: back-end structure, weight structure and front-end structure. The back-end structure is a matrix array $\mathcal{X}=(X_1,\cdots,X_d)$, where each $X_i$ is of size $M\times N$ and data type $T$. The weight structure $\mathcal{W}=I$ is the identity matrix of size $M\times M$ and data type $T$. The front-end structure $\mathcal{Y}=(Y_1,\cdots,Y_d)$ is also a matrix array, defined as:
\begin{equation}
\mathcal{Y}=\mathcal{W}\times \mathcal{X}=(\mathcal{W}\times X_1,\cdots,\mathcal{W}\times X_d)=(I\times X_1,\cdots,I\times X_d)=(X_1,\cdots,X_d)=\mathcal{X},
\end{equation}
where $Y_i=I \times X_i=X_i(1\leq i \leq d)$.

3) A reshaping connection transforms the back end into the front end with reshaping operation. It has three attributes: back-end structure, weight structure and front-end structure. The back-end structure is a matrix array $\mathcal{X}=(X_1,\cdots,X_d)$, where each $X_i$ is of size $M\times N$ and data type $T$. The weight structure is a $d$-channel vector array $\mathcal{W}=(W_1,\cdots,W_M)$, where $W_i=(0,\cdots,\underset{i}{\underbrace{1}},0,\cdots,0)(1\leq i \leq M)$ is an $M$-dimensional vector. The front-end structure $\mathcal{Y}$ is an $NMd$-dimensional vector, defined as:
\begin{equation}
\begin{aligned}
\mathcal{Y} & = (Y_1,\cdots,Y_{Md}) = (Y_{1\cdot1},Y_{2\cdot1},\cdots,Y_{M\cdot1},\cdots,Y_{i\cdot j},\cdots,Y_{1\cdot d},\cdots,Y_{M\cdot d}) \\
& = \mathcal{W}\times \mathcal{X} = \mathcal{W}\times (X_1,\cdots,X_d) = (\mathcal{W}\times X_1,\cdots,\mathcal{W}\times X_d) \\
& = ((W_1,\cdots,W_M)\times X_1,\cdots,(W_1,\cdots,W_M)\times X_d)\\
& = (W_1\times X_1,\cdots,W_M\times X_1,\cdots,W_1\times X_d,\cdots,W_M\times X_d),
\end{aligned}
\end{equation}
where $Y_{i\cdot j}=W_i\times X_j(1\leq i \leq M, 1 \leq j\leq d)$ is the $i$-th row of $X_j$, and it is an $N$-dimensional vector.

4) A full connection transforms the back end into the front end with matrix multiplication. It has three attributes: back-end structure, weight structure and front-end structure. The back-end structure is an $N$-dimensional vector $\mathcal{X}=X$. The weight structure $\mathcal{W}=W$ is a matrix of size $M\times N$ and data type $T$. The front-end structure $\mathcal{Y}=Y$ is an $M$-dimensional vector, defined as:
\begin{equation}
\mathcal{Y}=\mathcal{W}\times \mathcal{X}=Y=W\times X.
\end{equation}

\subsection{Standard graphical symbols for plain networks}

As a special case of capsule networks, plain networks can also be implemented by drawing them with standard graphical symbols, such as "data neuron", "ReLU neuron", "identical neuron", "arrow amplifying connection", and "amplifying connection". Some of these graphical symbols are shown in Figure 19(a-e).
\begin{figure}[htb]
  \centering
  \includegraphics[width=2.9in,height=0.5in]{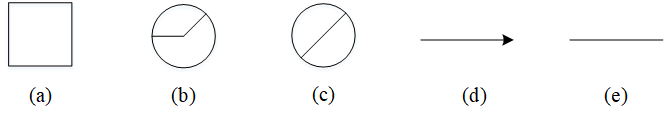}
  \caption{Graphical symbols for plain networks: (a) data neuron; (b) ReLU neuron; (c) identical neuron; (d) arrow amplifying connection; (e) amplifying connection.}
\end{figure}

A data neuron is a scalar input with data type $T$. A ReLU neuron transforms a scalar input $x$ into another scalar output $y$ with activation function ReLU, defined as:
\begin{equation}
y=ReLU(x).
\end{equation}

An identical neuron plays the identical function, namely,
\begin{equation}
y=I(x)=x.
\end{equation}

An (arrow) amplifying connection transforms the back end into the front end with amplification. It has three attributes: back-end scalar $x$, weight strength $w$, and front-end scalar $y$. The value of $y$ is computed by:
\begin{equation}
y=wx.
\end{equation}

In case of no ambiguity, amplifying connection can be in place of arrow amplifying connection.

\subsection{Graphical programming}

Based on graphical symbols of capsules and their connections, we can design a graphical programming platform to implement deep neural networks. For example, we can implement MLP and LeNet by drawing the capsule graphs in Figures 20-21.
\begin{figure}[htb]
  \centering
  \includegraphics[width=2.7in,height=0.5in]{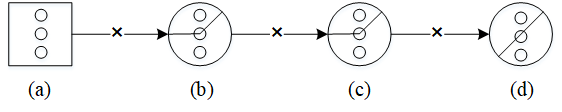}
  \caption{A graphical representation of the capsuled MLP, which is orderly composed of 1D-data capsule (a), 1D-ReLU capsule (b), 1D-ReLU capsule (c), and 1D-identical capsule (d).}
\end{figure}

\begin{figure}[htb]
  \centering
  \includegraphics[width=5.3in,height=0.5in]{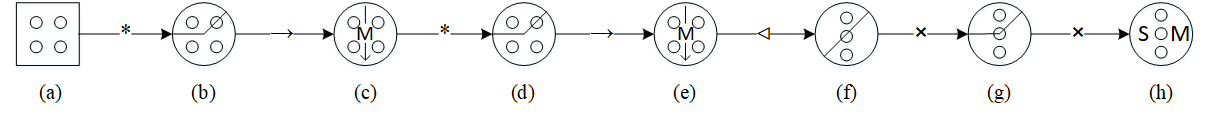}
  \caption{A graphical representation of the capsuled LeNet, which is orderly composed of 2D-data capsule (a), 2D-ReLU capsule (b), 2D-maximum downsampling capsule (c), 2D-ReLU capsule (d), 2D-maximum downsampling capsule (e), 1D-identical capsule (f), 1D-ReLU capsule (g), and 1D-softmax capsule (h).}
\end{figure}

As shown in Figure 20, the capsuled MLP consists of one 1D-data capsule, two 1D-ReLU capsules, and one 1D-identical capsule, together with three full connections. The dimensions of the capsules can be set to 2, 6, 4 and 2, with the same type of "float64". Accordingly, the full connection between module (a) and module (b) must have a back end with dimension=2 and data type = "float64", a weight matrix with height = 6, width = 2 and data type = "float64", and a front end with dimension = 6 and data type = "float64". The full connection between module (b) and module (c) must have a back end with dimension=6 and data type = "float64", a weight matrix with height =4, width =6 and data type = "float64", and a front end with dimension = 4 and data type = "float64". The full connection between module (c) and module (d) must have a back end with dimension=4 and data type = "float64", a weight matrix with height =2, width =4 and data type = "float64", and a front end with dimension = 2 and data type = "float64". Therefore, the capsuled MLP is equivalent to the plain MLP depicted in Figure 22.
\begin{figure}[htb]
  \centering
  \includegraphics[width=1.5in,height=1.5in]{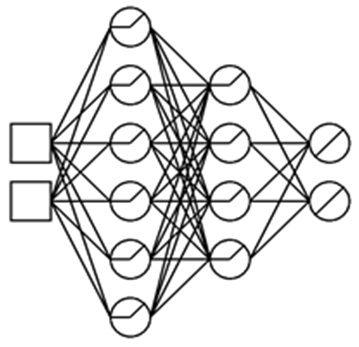}
  \caption{A graphical representation of the plain MLP.}
\end{figure}

As shown in Figure 21, the capsuled LeNet consists of one 2D-data capsule, two 2D-ReLU capsules, two 2D-maximum downsampling capsules, one 1D-identical capsule, one 1D-ReLU capsule and one 1D-softmax capsule, together with two convolutional connections, two transfer connections, one reshaping connection and two full connections. If the LeNet is applied to the MNIST dataset [50], module (a) can be a 2D-data capsule with height=28, weight=28, channel number=1, and data type = "float64". Module (b) can be a 2D-ReLU capsule with height = 24, width = 24, channel number = 32, and data type = "float64". The convolution connection between module (a) and module (b) can have a back end with height=28, weight=28, channel number=1, and data type = "float64", and a weight structure with kernel number= 32, height = 5, width = 5, channel number = 1, stride = 1, and data type = "float64", and a front end with height = 24, width = 24, channel number = 32, and data type = "float64". The other modules and their connections can be determined compatibly in a similar manner.
\begin{figure}[htb]
  \centering
  \includegraphics[width=5.5in,height=1.6in]{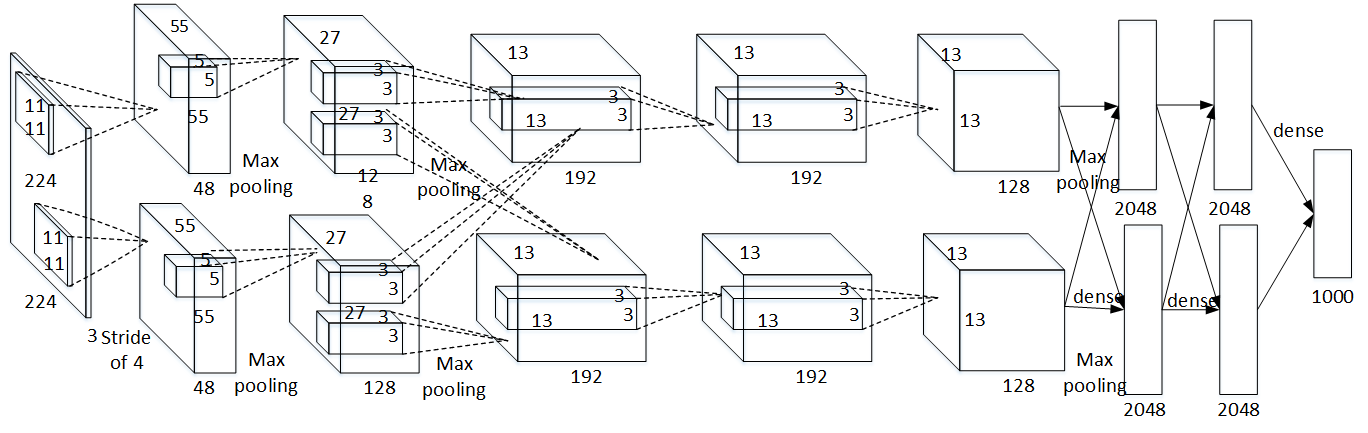}
  \caption{A graphical representation of AlexNet in tradition.}
\end{figure}

\begin{figure}[htb]
  \centering
  \includegraphics[width=5.5in,height=0.8in]{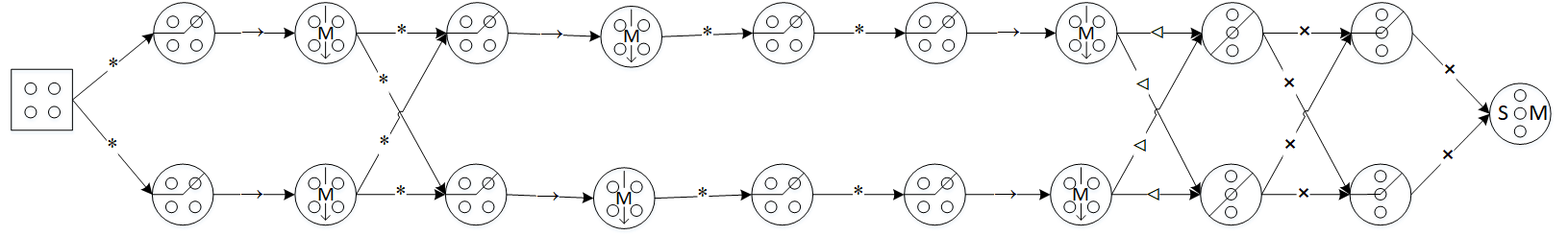}
  \caption{A graphical representation of AlexNet by standard symbols.}
\end{figure}

\begin{figure}[htb]
  \centering
  \includegraphics[width=3.5in,height=1.2in]{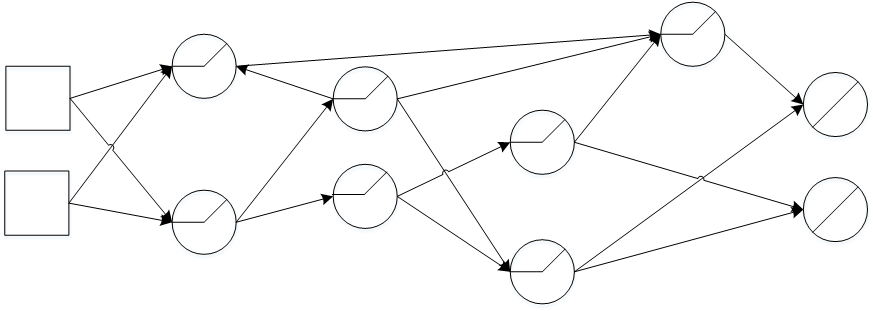}
  \caption{A deep neural network with flexible connections.}
\end{figure}

Like LeNet, we can implement AlexNet (see Figure 23) by drawing its capsule graph displayed in Figure 24. Using more graphical symbols, we can similarly implement VGGNet, GoogLeNet, ResNet, and so on. Moreover, we can even implement a deep neural network with flexible connections in Figure 25, and a deep capsule network with flexible connections in Figure 26. Obviously, drawing is easier, simpler and more convenient than coding in popular deep learning platforms such as Caffe, TensorFlow, MXNet, and CNTK.
\begin{figure}[htb]
  \centering
  \includegraphics[width=5.5in,height=1in]{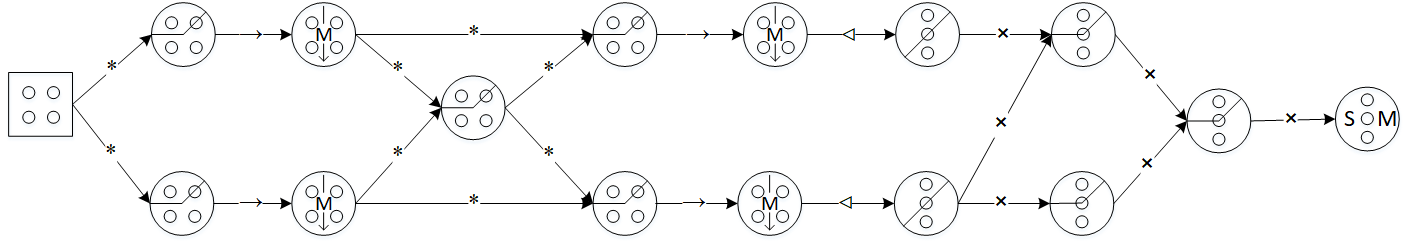}
  \caption{A deep capsule network with flexible connections.}
\end{figure}

\section{Conclusions}

We have used capsule networks to establish a unified framework of deep learning. We not only present a strict mathematical definition of neural networks, but also a generation theorem to clarify the intrinsic relationship between feedforward neural networks and connected directed acyclic graphs. As a generalization of neural networks, we obtain two types of capsule networks (i.e. layered and skip) to represent and extend various deep learning models. Additionally, we derive a universal backpropagation algorithm for capsule networks, and demonstrate its potential applications to graphical programming in combination with standard graphical symbols of plain networks and capsule networks. On a graphical programming platform, we can implement a variety of deep neural networks by drawing them directly, such as MLP, LeNet, AlexNet, and many others. Therefore, this kind of platform will make easier deep learning with beneficial complement to popular deep learning frameworks, such as Caffe, TensorFlow, MXNet and CNTK.

As future work, we will design a sufficient set of standard graphical symbols to describe deep neural networks, and then implement a graphical programming platform to make their construction more convenient and to facilitate their large-scale applications.

\section{Acknowledgements}

This work was supported by the National Natural Science Foundation of China under Grant 61876010.

\section*{References}

\medskip

\small

[1] Hinton, G. E.\ \& Salakhutdinov, R. R.\ (2006) Reducing the dimensionality of data with neural networks. {\it Nature} {\bf 313}(5786): 504-507.

[2] Krizhevsky, A., Sutskever, I.\ \& Hinton, G.E.\ (2012) Imagenet classification with deep convolutional neural networks. In F.\ Pereira, C.J.C.\ Burges, L.\ Bottou and K.Q.\ Weinberger (eds.), {\it Advances in neural information processing systems 25}, pp.\ 1097--1105. Cambridge, MA: MIT Press.

[3] Seide, F., Li, G.\ \& Yu, D.\ (2011) Conversational speech transcription using context-dependent deep neural networks. {\it Twelfth Annual Conference of the International Speech Communication Association}.

[4] Le, Q. V.\ (2013) Building high-level features using large scale unsupervised learning. {\it Acoustics, Speech and Signal Processing (ICASSP)}, 2013 IEEE International Conference on. IEEE, pp.\ 8595--8598.

[5] Gao, J., He, X.\ \& Yih, W.\ et al.\ (2014) Learning continuous phrase representations for translation modeling. {\it Proceedings of the 52nd Annual Meeting of the Association for Computational Linguistics (Volume 1: Long Papers)}, pp.\ {\bf 1}: 699-709.

[6] McCulloch, W. S\ \& Pitts, W.\ (1943) A logical calculus of the ideas immanent in nervous activity. {\it The bulletin of mathematical biophysics} {\bf 5}(4): 115-133.

[7] Hebb, D. O.\ (2006) The organization of behavior: A neuropsychological theory. {\it Psychology Press}.

[8] Rosenblatt, F.\ (1958) The perceptron: a probabilistic model for information storage and organization in the brain. {\it Psychological review}  {\bf 65}(6): 386.

[9] Minsky, M., Papert, S. A.\ \& Bottou, L.\ (2017) Perceptrons: An introduction to computational geometry. {\it MIT press}.

[10] Hopfield, J. J.\ (1982) Neural networks and physical systems with emergent collective computational abilities. {\it Proceedings of the national academy of sciences} {\bf 79}(8): 2554-2558.

[11] Ackley, D. H., Hinton, G. E.\ \& Sejnowski, T. J.\ (1985) A learning algorithm for Boltzmann machines. {\it Cognitive science} {\bf 9}(1): 147-169.

[12] Rumellhart, D.E.\ (1986) Learning internal representations by error propagation. {\it Parallel distributed processing: Explorations in the microstructure of cognition}  {\bf 1}:319-362.

[13] Werbos, P.\ (1974) Beyond regression: New tools for prediction and analysis in the behavior science. {\it Unpublished Doctoral Dissertation}  Harvard University.

[14] Cybenko, G.\ (1989) Approximation by superpositions of a sigmoidal function. {\it Mathematics of control, signals and systems} {\bf 2}(4): 303-314.

[15] Funahashi, K. I.\ (1989) On the approximate realization of continuous mappings by neural networks. {\it Neural networks} {\bf 2}(3): 183-192.

[16] Hochreiter, S.\ (1991) Untersuchungen zu dynamischen neuronalen Netzen. {\it Master's Thesis}, Institut Fur Informatik, Technische Universitat, Munchen.

[17] Hochreiter, S., Bengio, Y.\ \& Frasconi, P.\ et al.\ (2001) Gradient flow in recurrent nets: the difficulty of learning long-term dependencies. {\bf 28}(2):237-243.

[18] Hinton, G.E., Osindero, S.\ \& Teh, Y.W.\ (2006) A fast learning algorithm for deep belief nets. {\it Neural computation} {\bf 18}(7):1527-1554.

[19] Nagi, J., Ducatelle, F.\ \& Di Caro, G. A.\ et al.\ (2011) Max-pooling convolutional neural networks for vision-based hand gesture recognition. {\it Signal and Image Processing Applications (ICSIPA)}, 2011 IEEE International Conference on. IEEE, pp.\ 342--347.

[20] Wan, L., Zeiler, M.\ \& Zhang, S.\ et al.\ (2013) Regularization of neural networks using dropconnect. {\it International Conference on Machine Learning}, pp.\ 1058--1066.

[21] Hinton, G. E., Srivastava, N.\ \& Krizhevsky, A.\ et al.\ (2012) Improving neural networks by preventing co-adaptation of feature detectors. {\it Computer Science} {\bf 3}(4): 212-223.

[22] Ioffe, S.\ \& Szegedy, C.\ (2015) Batch normalization: accelerating deep network training by reducing internal covariate shift. {\it International Conference on International Conference on Machine Learning}, JMLR.org, pp.\ 448--456.

[23] Fukushima, K.\ (1979) Neural network model for a mechanism of pattern recognition unaffected by shift in position-Neocognitron. {\it IEICE Technical Report, A}, pp.\ {\bf 62}(10): 658-665.

[24] Fukushima, K.\ (1980) Neocognitron: A selorganizing neural network model for a mechanism of pattern recognition unaffected by shift in position. {\it Biological Cybernetics} {\bf 36}(4):193-202.

[25] LeCun, Y., Bottou, L.\ \& Bengio Y, et al.\ (1998) Gradient-based learning applied to document recognition. {\it Proceedings of the IEEE} {\bf 86}(11):2278-2324.

[26] Simonyan, K.\ \& Zisserman, A.\ (2014) Very Deep Convolutional Networks for Large-Scale Image Recognition. {\it Computer Science}.

[27] He, K. Zhang, X.\ \& Ren, S.\ et al.\ (2016) Deep residual learning for image recognition. {\it Proceedings of the IEEE conference on computer vision and pattern recognition}, pp.\ 770--778

[28] Russakovsky, O., Deng, J.\ \& Su, H.\ et al.\ (2015) Imagenet large scale visual recognition challenge. {\it International Journal of Computer Vision} {\bf 115}(3): 211-252.

[29] Szegedy, C. Liu, W.\ \& Jia, Y.\ et al.\ (2015) Going deeper with convolutions. {\it IEEE Conference on Computer Vision and Pattern Recognition}.

[30] Ding, C.\ \& Tao, D.\ (2015) Robust face recognition via multimodal deep face representation. {\it IEEE Transactions on Multimedia} {\bf 17}(11): 2049-2058.

[31] Schroff, F., Kalenichenko, D.\ \& Philbin, J.\ (2015) Facenet: A unified embedding for face recognition and clustering. {\it Proceedings of the IEEE conference on computer vision and pattern recognition}, pp.\ 815--823.

[32] Mnih, V., Kavukcuoglu, K.\ \& Silver, D.\ et al.\ (2015) Human-level control through deep reinforcement learning. {\it Nature} {\bf 518}(7540): 529.

[33] Silver, D., Schrittwieser, J.\ \& Simonyan, K.\ et al.\ (2017) Mastering the game of Go without human knowledge. {\it Nature} {\bf 550}(7676):354-359.

[34] Hinton, G., Deng, L.\ \& Yu, D.\ et al.\ (2012) Deep neural networks for acoustic modeling in speech recognition: The shared views of four research groups. {\it IEEE Signal Processing Magazine} {\bf 29}(6): 82-97.

[35] Iandola, F. N., Han, S.\ \& Moskewicz, M. W.\ et al.\ (2016) SqueezeNet: AlexNet-level accuracy with 50x fewer parameters and <0.5MB model size. {\it arXiv preprint arXiv:1602.0736}.

[36] He, K., Zhang, X.\ \& Ren, S.\ et al.\ (2014) Spatial pyramid pooling in deep convolutional networks for visual recognition. {\it European conference on computer vision}, Springer, Cham, pp.\ 346--361.

[37] Huang, G., Liu, Z.\ \& Weinberger, K.Q.\ et al.\ (2017) Densely connected convolutional networks. {\it Proceedings of the IEEE conference on computer vision and pattern recognition}, pp.\ {\bf 1}(2):3.

[38] Long, J., Shelhamer, E.\ \& Darrell, T.\ (2015) Fully convolutional networks for semantic segmentation. {\it Proceedings of the IEEE conference on computer vision and pattern recognition}, pp.\ 3431--3440.

[39] Ren, S., He, K.\ \& Girshick, R.\ et al.\ (2015) Faster r-cnn: Towards real-time object detection with region proposal networks. {\it Advances in neural information processing systems}, pp.\ 91--99.

[40] He, K., Gkioxari, G.\ \& Doll\'{a}r, P.\ et al.\ (2017) Mask r-cnn. {\it Computer Vision (ICCV), 2017 IEEE International Conference on. IEEE}, pp.\ 2980--2988.

[41] Redmon, J., Divvala, S.\ \& Girshick, R.\ et al.\ (2016) You only look once: Unified, real-time object detection. {\it Proceedings of the IEEE conference on computer vision and pattern recognition}, pp.\ 779--788.

[42] Liu, W., Anguelov, D.\ \& Erhan, D.\ et al.\ (2016) Ssd: Single shot multibox detector. {\it European conference on computer vision. Springer, Cham}, pp.\ 21--37.

[43] Ji, S., Xu, W.\ \& Yang, M.\ et al.\ (2013) 3D convolutional neural networks for human action recognition. {\it IEEE transactions on pattern analysis and machine intelligence} {\bf 35}(1): 221-231.

[44] Sabour, S., Frosst, N.\ \& Hinton, G.E.\ (2017) Dynamic routing between capsules. In I.\ Guyon, U.V.\ Luxburg, S.\ Bengio, H.\ Wallach, R.\ Fergus, S.\ Vishwanathan and R.\ Garnett (eds.), {\it Advances in Neural Information Processing Systems 30}, pp.\ 3859-3869. Cambridge, MA: MIT Press.

[45] Hinton, G. E., Sabour, S.\ \& Frosst, N.\ (2018) Matrix capsules with EM routing. {\it International Conference on Representation Learning}.

[46] Bauer, F. L.\ (1974) Computational Graphs and Rounding Error. {\it Siam Journal on Numerical Analysis} {\bf 11}(1): 87-96.

[47] Griewank, A.\ \& Walther, A.\ (2008) Evaluating derivatives: principles and techniques of algorithmic differentiation (2. ed.). {\it DBLP}.

[48] Li, Y.\ \& Shan, C.\ (2018) A Unified Framework of Deep Neural Networks by Capsules. {\it arXiv preprint arXiv:1805.03551}

[49] Ruder, S.\ (2016) An overview of gradient descent optimization algorithms. {\it arXiv preprint arXiv:1609.04747}.

[50] http://yann.lecun.com/exdb/mnist/.

\end{document}